\documentclass[11pt]{article}

\usepackage{authblk}
\usepackage{mathtools}
\usepackage{amsmath}
\usepackage{amsfonts}
\usepackage{listings}
\usepackage{graphicx}
\usepackage{longtable}
\usepackage[parfill]{parskip}
\usepackage{bbm}
\usepackage{float}
\usepackage{adjustbox}
\usepackage{algorithm}
\usepackage{algorithmic}
\usepackage{caption}
\usepackage{subcaption}
\usepackage{hyperref}
\usepackage{multirow}
\usepackage{rotating}
\usepackage{newtxtext, makeidx, multicol, footmisc}

\begin{document}

\author[1,2]{Pierre Miasnikof \thanks{corresponding author: p.miasnikof@mail.utoronto.ca}}
\author[1]{Mohammad Bagherbeik}
\author[1]{Ali Sheikholeslami}

\affil[1]{The Edward S. Rogers Sr. Dept. of Electrical \& Computer Engineering, University of Toronto, Toronto, ON, Canada}
\affil[2]{The University of Toronto Data Sciences Institute (DSI), Toronto, ON, Canada}

\title{Graph clustering with Boltzmann machines}
\date{}
\maketitle

\begin{abstract}
Graph clustering is the process of labelling nodes so that nodes sharing common labels form densely connected subgraphs with sparser connections to the remaining vertices. Because of its difficult formulation, we translate the intra-cluster density maximization problem to a distance minimization problem, through the use of a novel vertex-vertex distance that accurately reflects density.  Specifically, we extend the recent binary quadratic K-medoids formulation to graph clustering. We also generalize a quadratic formulation originally designed for partitioning complete graphs. Because binary quadratic optimization is an NP-hard problem, we obtain numerical solutions for these formulations through the use of two novel Boltzmann machine (meta-)heuristics. For benchmarking purposes, we compare solution quality and computational performances to those obtained using a commercial solver, Gurobi. We also compare clustering quality to the clusters obtained using the popular Louvain modularity maximization method. Our initial results clearly demonstrate the superiority of our problem formulations. They also establish the superiority of our Boltzmann machines over a traditional solver. In the case of smaller less complex graphs, Boltzmann machines provide the same solutions as Gurobi, but with solution times that are orders of magnitude lower. In the case of larger and more complex graphs, Gurobi either fails to return meaningful results within a reasonable time frame or returns inferior results. Finally, we also note that both our clustering formulations, the distance minimization and $K$-medoids, yield clusters of superior quality to those obtained with the Louvain algorithm.
\end{abstract}

\section{Introduction}

Graph clustering, also often called network community detection, is a pivotal task in the analysis of networks \cite{guideFortunato16}, data sets where the variables are represented as vertices on a graph with edges representing their interactions. In fact, it has even been described as {\it ``one of the most important and challenging problems in network analysis''}, in the recent literature \cite{LiudaSynthetic2019}. Graph clustering is an unsupervised learning task which consists of assigning each vertex of a graph to a cluster of arbitrary size. At its core, it is a combinatorial optimization problem. A successful clustering yields clusters that form dense induced subgraphs with sparse connections to vertices outside their respective clusters. Unfortunately, intra-cluster density maximization is a difficult problem to formulate and solve. For this reason, we introduce two heuristic approximations to it.

The work in this article builds upon the foundational framework of Fan and Pardalos \cite{FanPard2010LQ,FanPard2010CND}, Fan et al.~\cite{FanZhengPardalosIntervals2012} and Bauckhage et al.~\cite{Bauck2019}. The contributions of this article are a heuristic approximation of the intra-cluster density maximization problem through the adaptation of a binary graph partitioning formulation and the extension of the recently introduced quadratic $K$-medoids formulation to the case of graphs. We tailor these formulations through the use of a vertex-vertex distance that has been shown to accurately reflect density \cite{PMCplxNets2020,PMCplxNets2022}. Notably, both our reformulations offer a superior alternative to the leading Louvain method. To our knowledge, this article provides the first extension of the Fan and Pardalos graph partitioning formulations to the general case graph clustering problem. It also introduces the first application of the recent quadratic formulation of $K$-medoids to graph clustering. 

The quadratic graph partitioning formulation of Fan and Pardalos and Fan et al. was designed for the special cases of complete graphs or graphs where all (vertex) pairs distances were available. The $K$-medoids technique is a general purpose clustering technique that was not designed for or suited to graph data sets. It, too, requires distances between data points. To tailor these formulations to the general graph clustering problem, we use the Jaccard distance. Naturally, the choice of distance is of pivotal importance. For the specific purpose of clustering, it is primordially important to use a distance which reflects connectivity, not shortest path geodesics \cite{PMCplxNets2020,PMCplxNets2022}. 

We also use a novel clustering quality assessment. Instead of using the usual and problematic modularity function, we assess quality by examining intra- and inter-cluster densities. This novel approach has been shown to be superior \cite{PMEtAlWAW18,PMEtAlOUP20}. 

Another significant contribution is the numerical solution of both problem formulations using a Boltzmann machine (meta-)heuristic. For benchmarking purposes, solution times and quality are compared to those obtained with a leading commercial solver, Gurobi. We also conduct a comparison of the problem formulations, by examining the clustering quality yielded by the tremendously popular Louvain modularity maximization technique \cite{Louvain2008}. In all, we compare five different mathematical formulation--solver/solution technique combinations:
\begin{itemize}
\item Quadratic distance minimization solved using a Boltzmann machine,
\item Quadratic distance minimization solved using Gurobi,
\item Quadratic $K$-medoids formulation solved using a Boltzmann machine,
\item Quadratic $K$-medoids formulation solved using Gurobi and
\item Modularity maximization solved using the Louvain algorithm.
\end{itemize}

Numerical results highlight the superiority of our two mathematical formulations as graph clustering models. They also showcase the well-documented weaknesses of modularity as a clustering quality function and objective function to be maximized. Last but not least, they also outline the superiority of a Boltzmann (meta-)heuristic over a traditional leading-edge exact solver. 

The remainder of this article is organized as follows. After a brief survey of the literature, we provide a description of the graph clustering problem, from first principles. We then establish the link between our two mathematical programming clustering techniques and these defining principles.  

In closing, we also wish to call the readers' attention to the fact that this work only examines clustering of undirected (weighted or unweighted) graphs without multiple edges. Also, in our problem formulations, vertices are assigned to non-overlapping clusters. Cluster membership is assumed to be mutually exclusive.  

\section{Previous work} \label{litReview}
Graph clustering, also referred to as network community detection, is a distinct sub-field in unsupervised learning and clustering in particular \cite{ESL09}. The main distinction lies in the fact that graphs are not typically in metric space. Graphs are typically not represented in Euclidean space and all-pairs distances are not typically available, either. This difference makes most traditional clustering techniques, such as $K$-means for example \cite{ESL09}, inapplicable. Additionally, it should be noted that the very definition of graph clusters and graph clustering remains a topic of debate (e.g., \cite{GoodEtAl2010,Pitsoulis2018}). Nevertheless, virtually all authors agree that clusters (communities) are formed by sets of densely connected vertices that have sparser connections to the remaining vertices (e.g., \cite{Schaeffer2007,FortunatoLong2010,modWAW2016,EuroComb2017,PMEtAlWAW18,PMEtAlOUP20}).

A complete review of the graph clustering literature is beyond the scope of this article. For a very broad and thorough overview of the field, we refer the reader to the foundational work of Schaeffer \cite{Schaeffer2007}, Fortunato \cite{FortunatoLong2010} and Fortunato and Hric \cite{guideFortunato16}. Nevertheless, we note the existence of various competing graph clustering techniques, built on very different mathematical foundations. The main competing approaches to graph clustering are
\begin{itemize}
\item Spectral (e.g., \cite{Lux}),
\item Markov (e.g., \cite{MCMain}) and
\item Mathematical programming 
\begin{itemize}
\item Modularity maximization (e.g., \cite{Louvain2008,AloiseEtAl2012,GRASP_Pitsoulis,guideFortunato16})
\item Other objective functions (e.g., \cite{FanPard2010LQ,FanPard2010CND,LanciEtAlStat2011,FanZhengPardalosIntervals2012,PMEtAl8ICN,Pono2021}).
\end{itemize}
\end{itemize}
Modularity maximization is, by far, the most popular graph clustering formulation. The Louvain method is, by far, the most popular modularity maximization technique \cite{Louvain2008}. Its advantages are very short computation times, scalability and the fact it does not require the number of clusters as an input parameter.  Unfortunately, modularity also suffers from well-documented weaknesses (e.g., \cite{ResolLimitFortunato2007,AckerBD08,GoodEtAl2010,Kehagias2013,Pitsoulis2018,PMEtAlWAW18,PMEtAlOUP20}). In contrast, Fan and Pardalos \cite{FanPard2010LQ,FanPard2010CND} and Fan et al.~\cite{FanZhengPardalosIntervals2012} do not rely on modularity maximization. They exploit an all pairs distance (or similarity) between vertices and obtain clusters by minimizing (or maximizing) it. More recently, Bauckhage et al.~\cite{Bauck2019} introduce a quadratic unconstrained binary optimization (QUBO) formulation of the $K$-medoids technique \cite{PAM1990,ESL09}. $K$-medoids is not a typical graph clustering technique, because of its reliance on all pairs distances. We adapt both  the distance minimization and $K$-medoid formulations to the general graph clustering problem, through the use of Jaccard distances \cite{JaccOrig,Jacc71,Camby17,PMCplxNets2020,PMCplxNets2022}.  

Given our work includes a $K$-medoids formulation, it is important to highlight of the work of Ponomarenko et al. \cite{Pono2021}. Although their objective was to detect overlapping clusters in graphs, these authors adapted the original Partitioning Around Medoids algorithm of Kaufman and Rousseeuw \cite{PAM1990} to graphs, through the use of commute and amplified commute distances. The work in this article uses a different $K$-medoids formulation, the recently introduced quadratic unconstrained binary optimization (QUBO) formulation of Bauckhage et al. \cite{Bauck2019}. We also make use of a different distance metric, Jaccard distance, to adapt the $K$-medoids technique to graphs. 

Of course, it is important to also compare optimization-based approaches to other commonly used graph clustering techniques. Here, we must point out that spectral methods come with a heavy computational cost and do not work well on larger instances. This scale limitation was noted by Schaeffer \cite{Schaeffer2007}. Although some authors' more recent algorithms are described as ``faster and more accurate'', they still carry a heavy computational cost (e.g., \cite{FastScore2015}). More importantly, spectral methods have been described as ill-suited to sparse graphs \cite{guideFortunato16}. Unfortunately, clusterable graphs, graphs whose structure can be meaningfully summarized using clusters, are typically sparse. 

Markov-based techniques revolve around simulations of random walks over the graph. Such simulations require numerous matrix multiplications. Additionally, Markov clustering also requires various element-wise and row operations. An appealing feature of Markov clustering is that it does not require the number of clusters as a parameter input. While this feature may be advantageous in cases where a reasonable guess for the number of clusters is not known, in most cases domain knowledge does provide clues about this number. Since it is known that algorithms that do not require the number of clusters as an input parameter have been found to be less accurate than those that do require it \cite{guideFortunato16}, this initially appealing feature of Markov clustering may be a weakness, in many cases.

In contrast, optimization-based approaches lend themselves very well to approximate solution techniques, which carry a lower computational cost. Indeed, because of the typical NP-hardness of most graph clustering formulations \cite{Schaeffer2007,FortunatoLong2010}, solving these and other types of combinatorial optimization problems is often successfully done via (meta-)heuristic solution techniques (e.g., \cite{Papadimitriou98,Brownlee2011}), which explore subsets of the solution space. In the specific case of graph clustering, many authors have made use of (meta-)heuristic optimization techniques (e.g., \cite{AloiseEtAl2012,GRASP_Pitsoulis,GenClust14}), in order to circumvent the NP-hard nature of the problem and find approximate solutions. Additionally, (meta-)heuristic optimization techniques are easily parallelizable and well suited to implementation on high performance computing platforms.  

Numerous NP-hard problems have also been reformulated as Ising (QUBO) problems \cite{Fu86}. Such reformulations allow the implementation on massively parallel purpose-built hardware which yield solutions using simulated annealing \cite{IsingForm2014,Glover2018,Hahn2018Reducing,applic2019,physInspired2019,accel2020}. In fact, the graph partitioning problem is one of the original problems at the intersection of Ising modeling and the study of NP-complete problems \cite{IsingForm2014}.

\section{The graph clustering problem}
As stated previously, graph clustering is a central topic in the field of network science \cite{guideFortunato16,LiudaSynthetic2019}. Unfortunately, most graph clustering formulations lead to NP-hard problems \cite{Schaeffer2007,FortunatoLong2010}, which poses obvious computational challenges.

\begin{figure}
	\centering
	\begin{subfigure}{0.5\textwidth}
		\centering
		\includegraphics[width=\textwidth]{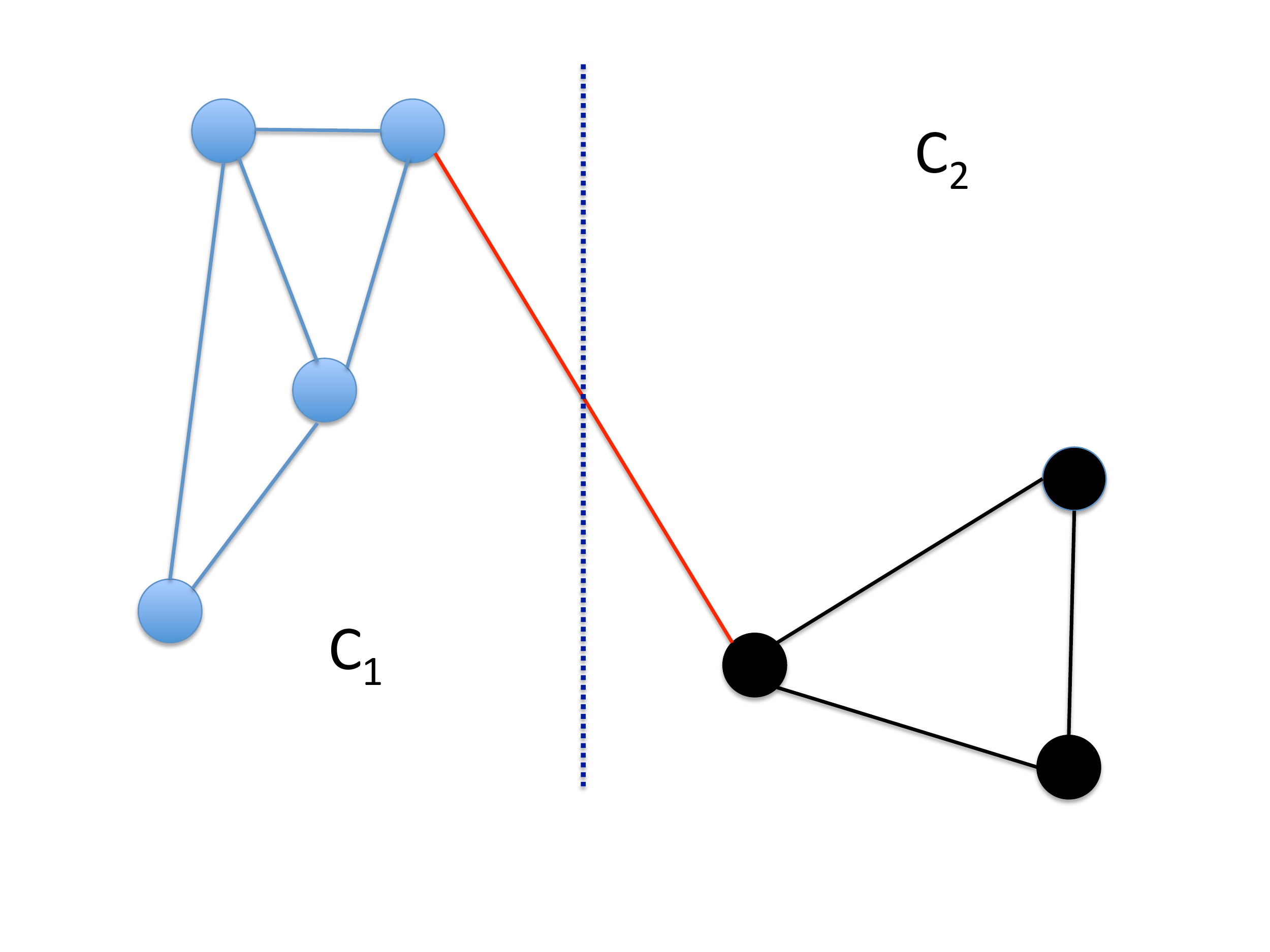}
		\caption{Well Clustered Graph} \label{Good1}
	\end{subfigure}%
	~ 
	\begin{subfigure}{0.5\textwidth}
		\centering
		\includegraphics[width=\textwidth]{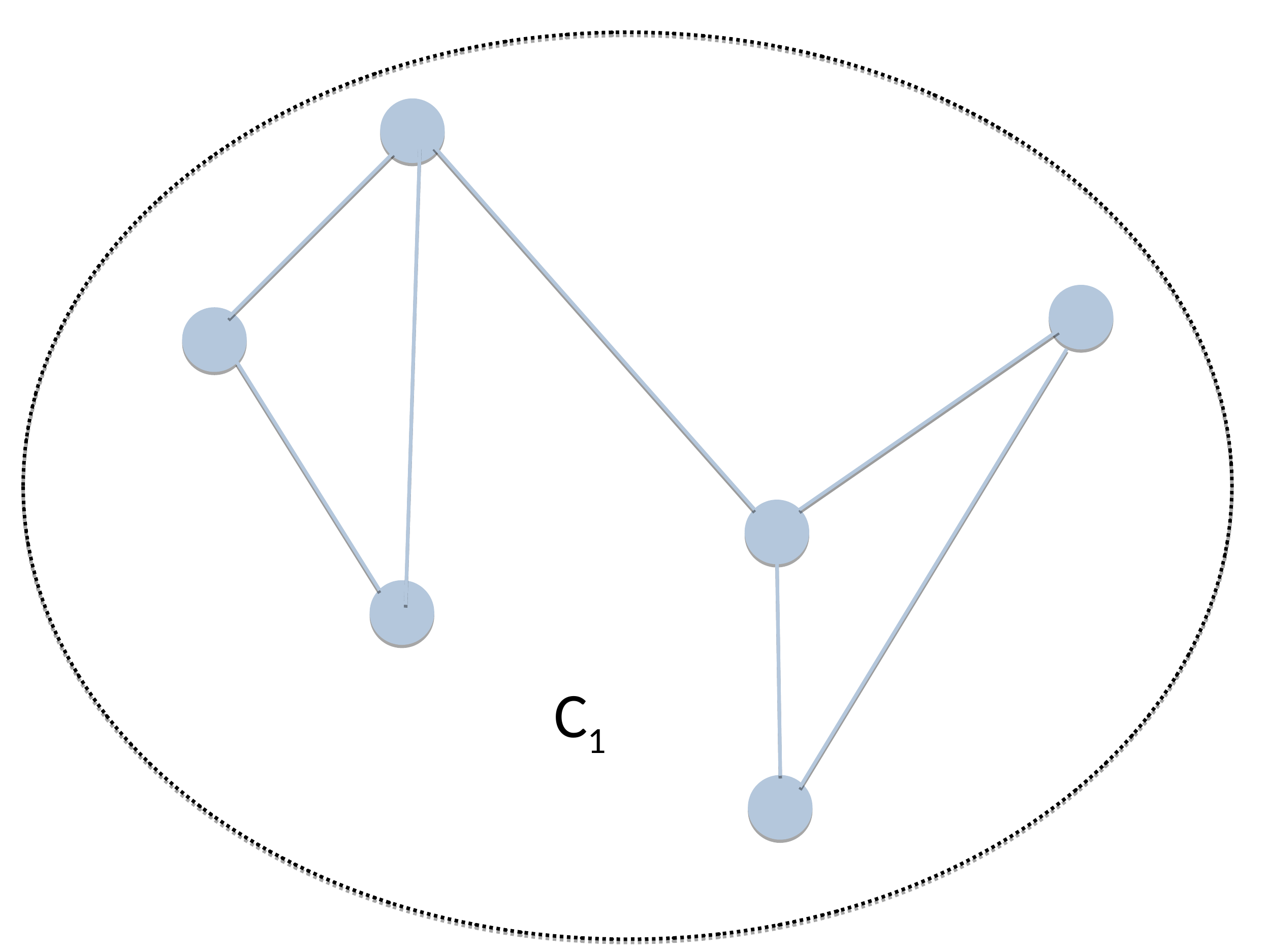}
		\caption{Improperly Clustered Graph} \label{NoGood1}
	\end{subfigure}
	\caption{Examples of Good and Bad Clustering}
	\label{goodNnogood}
\end{figure}

Graph clustering consists of grouping  vertices considered similar. Typically, similarity is defined by shared connections. Vertices that share more connections are defined as closer, more similar, to each other than to the ones with which they share fewer connections. Successful clustering results in vertices grouped into densely connected induced subgraphs (e.g., \cite{PMEtAlWAW18,PMEtAlOUP20}). Figure~\ref{goodNnogood} shows an example of a successful and an unsuccessful clustering.

Assessing output quality is another fundamental challenge of unsupervised learning. In the case of graph clustering, this challenge is greater due to the lack of a distance measure between data points (vertices). In accordance with the universally accepted understanding that graphs form dense subgraphs with sparser connection to remaining vertices, we use intra- and inter-cluster densities as benchmarks for clustering quality \cite{PMEtAlWAW18,PMEtAl8ICN,PMEtAlOUP20}.  

As mentioned earlier, in this work, we only consider clustering of undirected graphs without multiple edges. We also restrict our attention to non-overlapping clusters. In all our models, cluster membership is mutually exclusive.

\subsection{Clustering quality, intra- and inter-cluster densities} \label{Kappas}
While there is much debate surrounding the exact definition of a graph cluster (network community), the consensus view is that they are formed by densely connected sets of vertices that have sparser connections to the remaining graph. In accordance with this consensus, we use comparisons of intra-cluster, inter-cluster and overall densities as measures of clustering quality. Such comparisons have been shown to offer far superior assessments of clustering quality than the most popular modularity quality functions \cite{PMEtAlWAW18,PMEtAlOUP20}. These quantities are defined below.
\begin{eqnarray*}
\text{Graph (overall) density: }\mathcal{K} &=& \frac{\vert E \vert}{0.5 \times N (N-1)} \\
\text{Intra-cluster density (cluster $i$) : } \mathcal{K}_{\text{intra}}^{(i)} &=& \frac{\vert e_{ii} \vert}{0.5 \times n_i (n_i-1)} \\
\text{Inter-cluster density (cluster pair $i,j$) : } \mathcal{K}_{\text{inter}}^{(ij)}&=& \frac{ \vert e_{ij} \vert }{ n_i \times n_j  } \\
\end{eqnarray*}
In these definitions above, the variables used are
\begin{itemize}
\item $\vert E \vert$: the total number of edges,
\item $N \; (= \vert V \vert)$: the total number of vertices,
\item $\vert e_{ii} \vert$: the total number of edges connecting vertices in cluster $i$,
\item $\vert e_{ij} \vert$: the total number of edges connecting vertices in clusters $i$ and $j$ and
\item $n_i$: the total number of vertices in cluster $i$.
\end{itemize}

To gain a graph-wide and probabilistic view, we begin by noting that a graph's density, $\mathcal{K}$, can be interpreted as the Bernoulli probability that two arbitrary vertices are connected by an edge. Similarly, for each cluster $i$ or cluster pair ($i,j$), the densities  $\mathcal{K}_{\text{intra}}^{(i)}$ and $\mathcal{K}_{\text{inter}}^{(ij)}$ can be interpreted as the empirical estimate of intra-cluster and inter-cluster edge probability. These quantities are estimates of the probaility that two vertices in cluster $i$ and two vertices, one in cluster $i$ the other in cluster $j$, are connected by an edge. For a graph-wide view of these probabilities that is not sensitive to cluster size, that does not suffer from the well known resolution limit (e.g., \cite{ResolLimitFortunato2007}), we take the sample means. These means are the empirical estimates of the intra-/inter-cluster Bernoulli edge probabilities. We denote estimates, as opposed to actual probabilities, as $\hat{P}$. (In the definitions below, $C$ is the total number of clusters and $c_i$ is the cluster to which vertex $i$ has been assigned.) 
\begin{eqnarray*}
\mathcal{K} &=& \frac{\vert E \vert}{0.5 \times N (N-1)} = \hat{P}(e_{ij}) \\
\bar{\mathcal{K}}_{\text{intra}} &=& \frac{1}{C} \sum_i^C \mathcal{K}_{\text{intra}}^{(i)} = \hat{P}(e_{ij} \vert c_i = c_j) \\
\bar{\mathcal{K}}_{\text{inter}} &=& \frac{1}{0.5 \times C (C-1)} \sum_i^C \sum_{j=i+1}^C \mathcal{K}_{\text{inter}}^{(ij)} = \hat{P}(e_{ij} \vert c_i \ne c_j) \\
\end{eqnarray*}

Through these quantities, we can describe the properties of a good clustering and measure clustering quality. Indeed, in a well-clustered graph, we expect clusters, on average, to form dense induced subgraphs. We also expect cluster pairs, on average, to form sparse bi-partite graphs (when ignoring intra-cluster edges). In summary, in the case of a good clustering, the inequalities below must hold. 
\[
\bar{\mathcal{K}}_{\text{inter}} < \mathcal{K} < \bar{\mathcal{K}}_{\text{intra}}
\]

\begin{figure}
	\begin{subfigure}{0.3\textwidth}
	\centering
	\includegraphics[width=\textwidth]{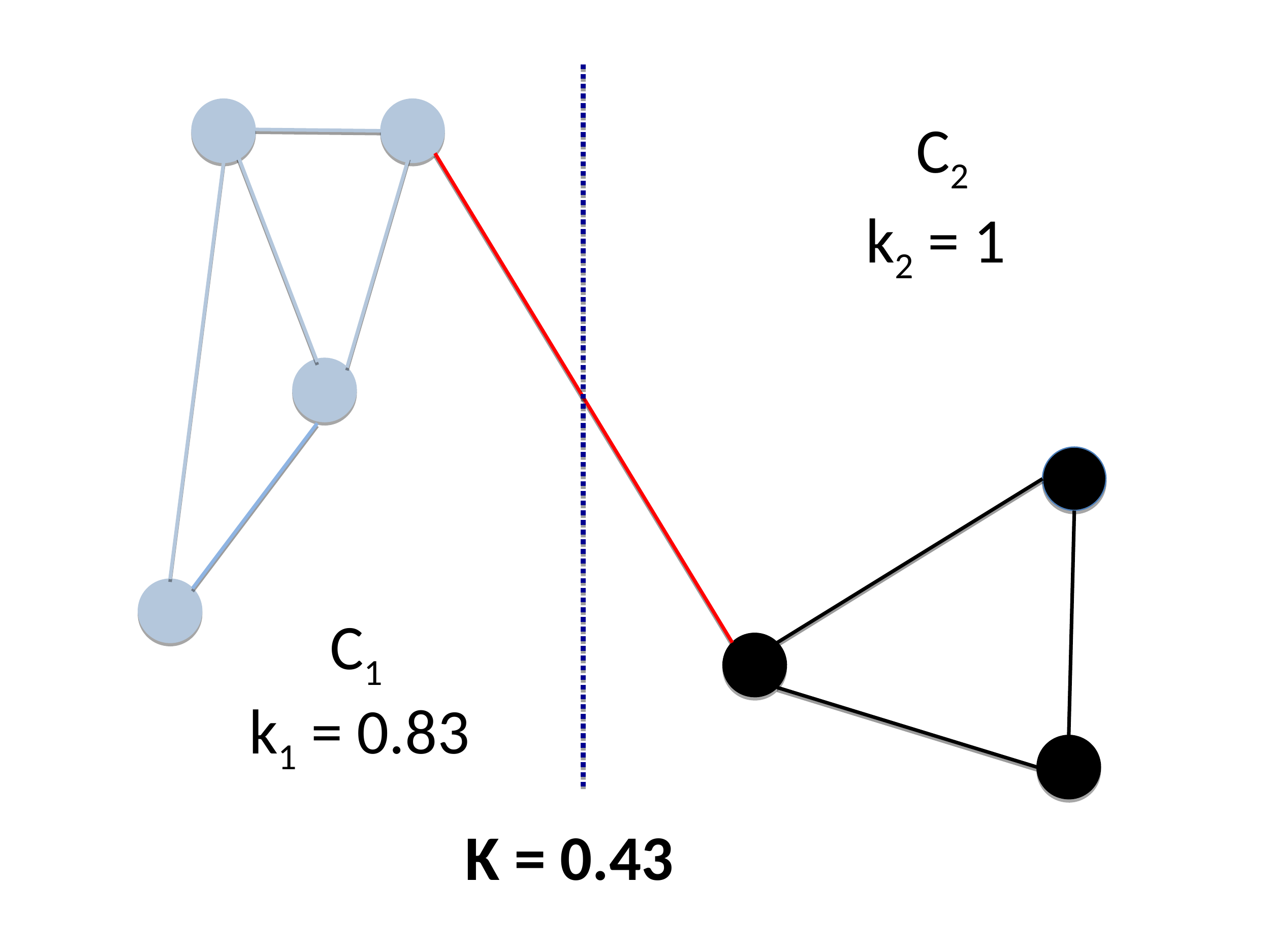}
	\caption{High intra-cluster and low inter-cluster densities} \label{exGood}
\end{subfigure}%
~ 
\begin{subfigure}{0.3\textwidth}
	\centering
	\includegraphics[width=\textwidth]{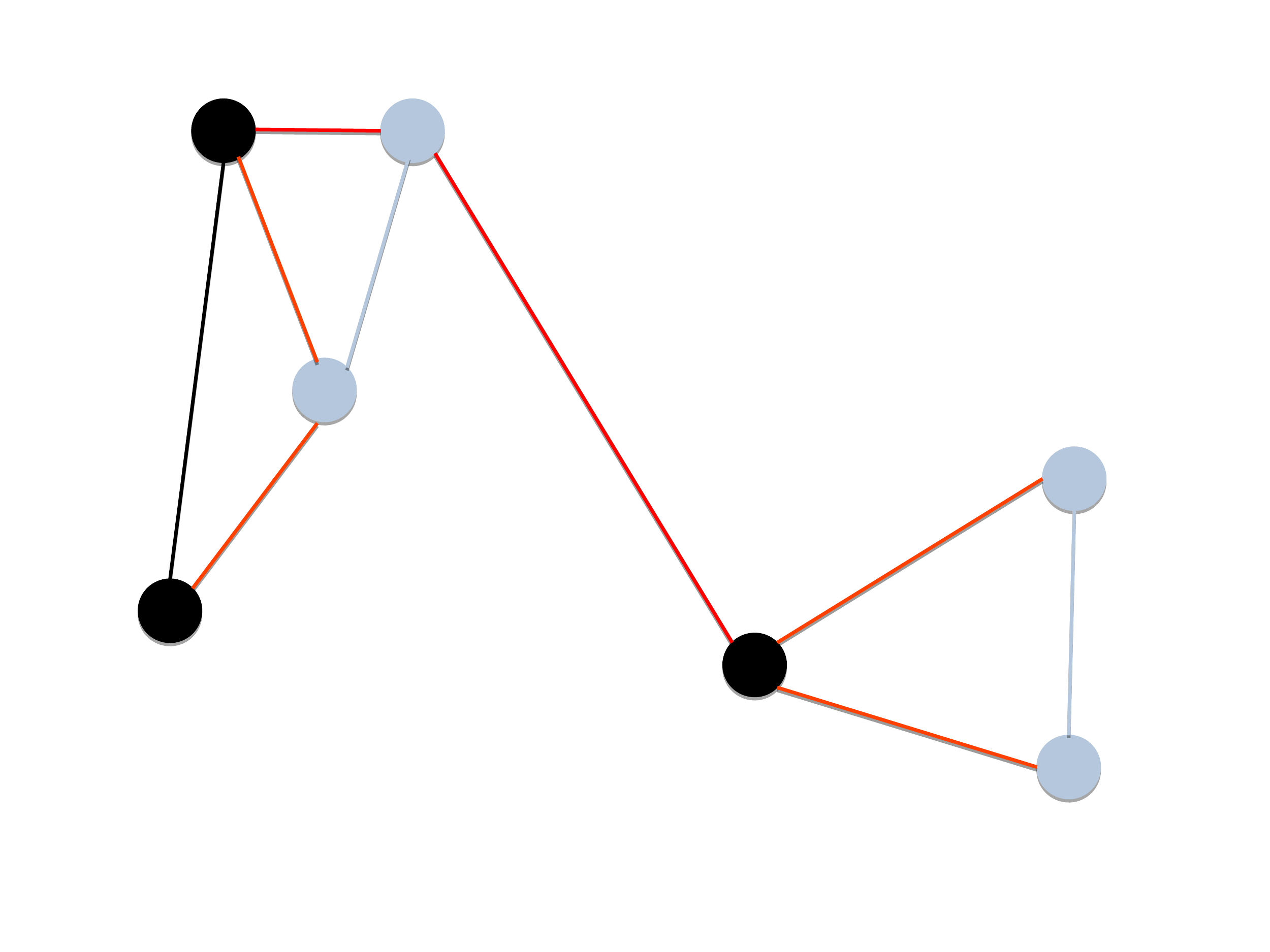}
	\caption{Improperly Clustered Graph} \label{exNoGood1}
\end{subfigure}
~
\begin{subfigure}{0.3\textwidth}
	\centering
	\includegraphics[width=\textwidth]{badClusters.pdf}
	\caption{Improperly Clustered Graph} \label{exNoGood2}
\end{subfigure}
\caption{Densities of Good and Bad Clustering}
\end{figure}

Figure~\ref{exGood} illustrates our density-based quality assessment with an example of good clustering. In this example of an arguably very well-clustered graph, the intra-cluster density of cluster $c_1$ (blue vertices) is $\mathcal{K}_{\text{intra}}^{(1)} = 0.83$. The intra-cluster density of cluster $c_2$ (black vertices) is $\mathcal{K}_{\text{intra}}^{(2)} = 1$. The mean intra-cluster density is $\bar{\mathcal{K}}_{\text{intra}} = \frac{1}{2}(0.83 + 1) = 0.92$. The mean inter-cluster density is $\bar{\mathcal{K}}_{\text{inter}} = \frac{1}{4 \times 3} = 0.08$ (there is only one cluster pair) and the graph's density is $\mathcal{K} = 0.43$. Consequently, the inequality $\bar{\mathcal{K}}_{\text{inter}} =  0.08 < \mathcal{K} = 0.43 < \bar{\mathcal{K}}_{\text{intra}} = 0.92$ holds. 

In contrast, Figure~\ref{exNoGood1} illustrates our quality metrics with an example of bad clustering. Both clusters have an intra-cluster density of $\frac{1}{3}$, for a $\bar{\mathcal{K}}_{\text{inter}} = \frac{1}{3}$. Inter-cluster density (inter-cluster edges are in red) is $\bar{\mathcal{K}}_{\text{inter}} = \frac{1}{2}$ (only one pair of clusters here too). Of course, graph density remains the same as in the previous case, $\mathcal{K} = 0.43$. Here, the inequalities observed in the previous example are reversed $\bar{\mathcal{K}}_{\text{inter}} = \frac{1}{2} > \mathcal{K} = 0.43 > \bar{\mathcal{K}}_{\text{intra}} = \frac{1}{3}$. In the case of the poorly clustered graph in Figure~\ref{exNoGood2}, the inequalities do not hold either. In that arguably degenerate case, all vertices are clustered in the same cluster. As a result, mean intra-cluster density is equal to the graph's density. Also as a result, there are no inter-clusters edges, so mean inter-cluster density is $0$. In summary, the necessary inequalities for a good clustering do not hold, instead we have the following: $\bar{\mathcal{K}}_{\text{inter}} = 0 < \mathcal{K} = \bar{\mathcal{K}}_{\text{intra}} =  0.43 $.

\section{Mathematical formulations} \label{mathform}
Good clusters form dense induced subgraphs. Correspondingly, an optimization-based clustering formulation should consist of assigning vertices to clusters such that intra-cluster density is maximized. This problem formulation consists of maximizing the objective function $f_o$ shown below. 
\begin{eqnarray*}
	f_o &=& \sum_k \frac{ \vert e_{kk} \vert}{0.5 \times n_k (n_k-1)} \\
	&=& \sum_k \frac{\sum_i \sum_{j > i} x_{ik} x_{jk} w_{ij}}{0.5 \times \left( \sum_i x_{ik} \right) \left( \left( \sum_i x_{ik}\right) -1 \right)} \; .
\end{eqnarray*}
In this formulation, we use the following variables,
\begin{itemize}
	\item $e_{kk}$: the set of edges connecting two vertices in cluster $k$,
	\item $n_k$: the number of vertices in cluster $k$
	\item and $x_{ik} \in \{0,1\}$: the (binary) decision variable which takes the value of 1 if vertex i is assigned to cluster $k$ ($0$ otherwise).
\end{itemize}
Unfortunately,  formulation is fractional and can only be solved by iterative algorithms \cite{PMEtAl8ICN}. Fortunately, there exist alternative mathematical programming based graph clustering formulations. In this article, we customize, implement and test a binary quadratic distance minimization formulation \cite{FanPard2010LQ,FanPard2010CND,FanZhengPardalosIntervals2012} and a quadratic $K$-medoids formulation \cite{Bauck2019}. We use these two models as (heuristic) approximations to the intra-cluster density maximization problem and validate the quality of the clustering results by examining intra- and inter-cluster densities.

\subsection{Binary quadratic formulation distance minimization (QP)} \label{FPQuad}
We begin with the binary quadratic formulation of Fan and Pardalos \cite{FanPard2010LQ,FanPard2010CND} and Fan et al.~\cite{FanZhengPardalosIntervals2012}. (Note: Fan and Pardalos have also done work on cut-based graph clustering \cite{FanPard2012}. This work is not considered in this study.) Their formulation presented both a distance minimization and an equivalent similarity maximization problem. In the first case (minimization), the parameter $d_{ij}$ is a distance separating vertices $i$ and $j$. In the second case, that parameter represents similarity. In both cases, $x_{ik}$ is a binary variable that takes the value of $1$ if vertex $i$ is assigned to cluster $k$. A constraint ensures each vertex is assigned to exactly one cluster.  The full minimization model we use in our experiments is presented below. 
\begin{equation*}
\begin{array}{rrclcl}
\displaystyle \min_{x} & \multicolumn{3}{l}{\sum_i \sum_{j>i} \sum_k x_{ik}x_{jk} d_{ij}} \\
\textrm{s.t.} & \sum_k x_{ik} & = & 1 & &\forall i\\
& x_{ik} & \in & \{0,1\} & & \forall i,k \\
\end{array}
\end{equation*}
As initially presented, this model cannot be directly applied to most graphs. Indeed, in most cases, all pairs distance or similarity matrices are not available. To generalize this model to the case of typical graphs, we need a distance or similarity that reflects connectivity between vertices. We require a distance whose pairwise minimization will lead to densely connected clusters. Shortest path distances do not have this property. Instead, we use the Jaccard distance \cite{JaccOrig,Jacc71,Camby17} between each vertex pair for the parameters $d_{ij}$. These distances can be obtained directly from the graphs' adjacency matrices. Full details, including a discussion on the inverse relation between Jaccard distance and intra-cluster density, can be found in Miasnikof et al. (2021,2022) \cite{PMCplxNets2020,PMCplxNets2022}. A short description is also provided in Section~\ref{JDist}.

\subsection{Quadratic $K$-medoids formulation ($K$-med)} \label{Kmed}
$K$-medoids is a clustering technique that selects $K$ exemplars (medoids) from the data set that will form the central point of each cluster. The remaining points are then assigned to the nearest medoid, thus forming $K$ clusters \cite{PAM1990,ESL09}. The $K$-medoids problem is NP-hard and it is typically solved using iterative algorithm heuristics \cite{Lloyd82,PAM1990,ESL09}. In late 2019, Bauckhage et al.~\cite{Bauck2019} presented a mathematical programming formulation for $K$-medoids. Their formulation is in the form of a quadratic unconstrained binary optimization (QUBO) problem, in order to take advantage of the newly available purpose built architectures for solving this type of (NP-hard) optimization problems  \cite{IsingForm2014,Glover2018,Hahn2018Reducing,applic2019,physInspired2019,accel2020}.

In the formulations that follow, $z_i = 1$ if the data point $i$ is selected as an exemplar and $z_i = 0$ otherwise. The vector $\vec{1}$ is a vector of ones of appropriate dimension. The number of data points to be clustered is given by $n$. The matrix $\Delta$ is an $(n \times n)$ matrix containing the distances separating all pairs of points. The distance separating each pair of points $i,j$ is  denoted $d_{ij}$. In mathematical form, we have
\begin{eqnarray*}
	\vec{z} &=& [z_1 z_2 \ldots z_n]^T, \; \forall z_i \in \{0,1\}, \\
	\vec{1} &=& [1 1 \ldots 1]^T \in \mathbb{R}^n \text{ and } \\
	\underset{(n \times n)}{\Delta} &=& [d_{ij}] \, .
\end{eqnarray*}

The original problem presented by Bauckhage et al.~\cite{Bauck2019} was formulated to minimize the distance between each exemplar and the remaining data points (maximize centrality), while maximizing the distance between each of these exemplars (maximize scattering). This trade-off optimization was achieved by the inclusions of the non-negative trade-off parameters $\alpha \text{ and }\beta$. Because it was in QUBO form, the objective function also included a non-negative penalty coefficient $\gamma$, which was applied to the constraint. The complete formulation can be expressed as:
\[
 \min \left\{ f_o = \underbrace{\beta \vec{z}^T\Delta\vec{1}}_{\text{centrality}} - \underbrace{\alpha\frac{1}{2}\vec{z}^T\Delta \vec{z}}_{\text{scattering}} 
 + \underbrace{\gamma (\vec{z}^T \vec{1} - k)^2}_{\text{constraint}} \right\} \, .
\]
In its original presentation, this formulation was not aimed at or suited to the graph clustering problem. We tailor it to graph clustering and Boltzmann machines, in two ways. First, we remove the quadratic penalty constraint, which is unnecessary with our Boltzmann heuristic. The built-in constraint handling offered by the $K$-hot encoding of our Boltzmann machine also reduces the burden of parameter-tuning, by eliminating one of the three parameters in the original Bauckhage et al. formulation \cite{Bauck2019}. Full details of the Boltzmann heuristic are presented in Section~\ref{Bheuristics}.

Second, as in the case of the quadratic distance minimization problem, we use the Jaccard distances \cite{JaccOrig,Jacc71,Camby17,PMCplxNets2020,PMCplxNets2022} as a distance metric $d_{ij}$. These modifications allow the application of the $K$-medoids clustering technique to the general case of graphs, where all pairs distances are not available. Finally, we also test the ``robust-ification'' of distances using a Welsch’s M-estimator, as suggested by Bauckhage et al. \cite{Bauck2019}, but do not find it useful. Through trial, we conclude the unmodified Jaccard distance yields better results.

Our unconstrained formulation provides a completely equivalent problem (to the original Bauckhage et al. problem). The cardinality constraint for exemplars ($\vec{z}^T\vec{1} = K$) is enforced directly by the $K$-hot encoding of the Boltzmann machine. The final problem can be expressed as:
\begin{equation*}
\begin{aligned}
& \min && \left\{ f_o = \underbrace{\beta \vec{z}^T\Delta\vec{1}}_{\text{centrality}} - \underbrace{\alpha\frac{1}{2}\vec{z}^T\Delta \vec{z}}_{\text{scattering}} \right\}\\
& \text{s.t} && \vec{z}^T\vec{1} = K \quad \text{(constraint no longer in objective)}\\
&&& z_i \in \{0,1\}, \quad \forall i \in V \, .
\end{aligned}
\end{equation*}

Finally, we set the trade-off parameters to $\alpha = 2$ and $\beta = 1.05 \times \frac{K + 1}{n}$ (recall $K$ is the number of exemplars, $n$ is the number of points, vertices in this case). Through trial and error, we find these parameters provide good results. Naturally, the tuning of these two parameters must be analyzed more closely. It should become the focus of a future investigation.

\subsection{Jaccard Distance} \label{JDist}
The Jaccard distance separating two vertices $i$ and $j$ is defined as
\[
d_{ij} = 1 - \frac{ \vert c_i \cap c_j \vert  }{ \vert c_i \cup c_j \vert} \in [0,1] \, .
\]
Here, $c_i \, (c_j)$ represents the set of all vertices with which vertex $i \, (j)$ shares an edge. This distance measures the similarity in the respective neighborhoods of two nodes. As stated earlier, this quantity has been shown to be inversely related to intra-cluster cluster density \cite{PMCplxNets2020,PMCplxNets2022}. 

\subsection{Louvain: modularity maximization}
Because of its widespread use and its status as the ``state of the art'' graph clustering technique, we compare the clustering quality obtained with our two formulations to that obtained with the Louvain algorithm \cite{Louvain2008}. The Louvain algorithm is a greedy iterative  heuristic technique that maximizes a clustering quality function known as modularity \cite{NewGirvOrig2004}. Unlike our two formulations presented in this article, the Louvain technique does not require the number of clusters as an input parameter. 

The Louvain method is known to be very fast, we do not include it in our study to provide a comparison of solution times. Rather, we want to compare solution qualities. Our goal is to assess the validity of our two mathematical formulations, distance minimization and $K$-medoids, by comparing their output to that obtained by maximizing modularity.

Below, we present modularity $(Q)$, as shown in Blondel et al.~\cite{Louvain2008}. Modularity is defined as 
\[
	Q = \frac{1}{2m} \sum_{i,j} \left[ A_{ij} - \frac{k_i k_j}{2m} \right] \delta (c_i,c_j) \, .
\]
In the equation above, 
\begin{itemize}
\item $m= \vert E \vert $ is the total number of edges in the graph, 
\item $A_{ij}$ is the element at the intersection of the $i$-th row and $j$-th column of the adjacency matrix, 
\item $k_i = \sum_j A_{ij}$ is the degree of vertex $i$,
\item $\delta(c_i,c_j)$ is the Kroenecker delta function, it is equal to one if $c_i = c_j$ and zero otherwise and
\item $c_i$ is the cluster to which vertex $i$ is assigned by the algorithm.
\end{itemize} 
Modularity always lies on the interval $[-\frac{1}{2},1]$ \cite{modBrandes2007}. Values greater than $0.3$ typically indicate a significant clustering \cite{ModClausetNewman2004}.

A full discussion of modularity or the Louvain algorithm are beyond the scope of this article. Results obtained with the Louvain algorithm are simply used for the purpose of comparison. For a very thorough discussion of these topics we refer the reader to the original authors \cite{Louvain2008}, to the work of Fortunato \cite{FortunatoLong2010} and Fortunato and Hric \cite{guideFortunato16}. As highlighted earlier, the limitations of modularity (and Louvain consequently) are also well described in the literature (e.g., \cite{ResolLimitFortunato2007,AckerBD08,GoodEtAl2010,Kehagias2013,Pitsoulis2018,PMEtAlWAW18,PMEtAlOUP20}).

\section{Boltzmann machines} \label{Bheuristics}
Boltzmann machines (BM) are neural networks that have been used to heuristically solve combinatorial optimization problems, for some time now \cite{Boltz1989,Boltz2001}. These machines encode an optimization problem into a graph-like structure where each decision variable is represented by a node. These nodes are logical units which can be inactive (set to $0$) or activated (set to $1$). The original objective function is then encoded as an energy function, using these logical units as decision variables and the edge weights as coefficients. Simulated annealing is used to minimize the energy function. 

To avoid confusion with the original graph being clustered, we only use the term ``{\it graph}'' to refer to the original graph being clustered. The graph-like Boltzmann encoding is referred to as the ``{\it Boltzmann network}'' or simply ``{\it network}''. The term ``{\it vertex}'' always refers to the original graph's vertices, while the Boltzmann network's nodes are always referred to as ``{\it units}''. We apply this terminological convention throughout the remainder of this article.

The advantage of the Boltzmann encoding is that it allows large-scale parallelization \cite{Boltz1989}. We further enhance this parallelization through the application of a parallel tempering scheme \cite{SwendsenRepl86,Ptemp1991,HukushimaExch96,Ptemp2005,Ptemp2020}. This temperature exchange scheme allows for a better coverage of the solution space. We use a multi-threaded implementation of the Boltzmann machine with parallel tempering, on a 64-core machine with a single instruction multiple data scheme (SIMD). 

Two different variations of the Boltzmann machine were created. Each structure implicitly captures the cluster membership or the medoid cardinality constraints of the mathematical formulations in Section~\ref{mathform}. Unlike with digital or quantum annealers, a QUBO objective function which encapsulates constraints is not required \cite{Glover2018}. However, each structure encodes the search space and objective function differently. 

\subsection{One-hot encoded machine}
Our first variation of the Boltzmann machine is a one-hot encoded machine. We call this version the integer Boltzmann machine \cite{BaghBoltz2020}. In this architecture, each one of the $N$ vertices of the original graph to be clustered is represented by $K$ Boltzmann machine units. For a given vertex, each of the $K$ corresponding units represents the cluster membership to one of the $K$ clusters. In total, we have $N \times K$ units. Under this architecture, by design, exactly one of the $K$ units representing a given vertex is forced to take a value of one. Therefore, our cluster membership constraint is always enforced structurally. We apply this variation of the Boltzmann machine to the quadratic distance minimization formulation (QP) in Section~\ref{FPQuad}. 

The energy function $(E)$ to be minimized is identical to the problem formulation presented in Section~\ref{FPQuad}:
\[
E(\vec{x}) = \sum_i \sum_{j>i} \sum_k x_{ik}x_{jk} d_{ij} \, .
\] 
Each decision variable $x_{ik}$ represents a Boltzmann network logical unit. The vector $\vec{x}$ is a vector with scalar components $x_{ik}$. If vertex $i$ is assigned to cluster $k$ then the corresponding unit $x_{ik}$ is activated.

\subsection{$K$-hot encoded machine}
Our second variation is a $K$-hot encoded machine \cite{BaghKHot2021}. We apply this variation of the Boltzmann machine to the $K$-medoids formulation ($K$-med). Here, each of the $N$ vertices is represented by only one unit, indicating whether or not it is an exemplar. Once again, a structural constraint ensures exactly $K$ out of the $N$ units have a value of one, at all times. Naturally, the $K$-hot encoding has a lower memory requirement than the one-hot encoding machine and most solver architectures. 

The energy function $(E)$ to be minimized is identical to the problem formulation presented in Section~\ref{Kmed},
\[
E(\vec{x}) =  \underbrace{\beta \sum_i x_i\left(\sum_{j \ne i} d_{ij} \right)}_{\text{centrality}} - \underbrace{\alpha \sum_i \sum_{j > i} x_i x_j d_{ij}}_{\text{squattering}} \, .
\]
In this case, the decision variables $x_i$ represent the BM unit corresponding to the $i$-th graph vertex. The vector $\vec{x}$ is a vector with scalar components $x_i$. If vertex $i$ is chosen as an exemplar, the unit $x_i$ is activated.

\subsection{Simulated annealing and parallel tempering}
Typically, Boltzmann machines are combined with simulated annealing to find the optimal machine state. At its core, simulated annealing is a random walk (Markov chain) \cite{SAOrig1983} through the search space. In both our variations of the Boltzmann machine, we use the Metropolis-Hastings algorithm to accept or reject solutions. 

Unfortunately, simulated annealing often fails to converge to a global optimum and remains stuck in a local one instead. This failure is mainly due to monotonically decreasing temperatures used in the Metropolis-Hastings algorithm. Parallel tempering mitigates this weakness \cite{SwendsenRepl86,Ptemp1991,HukushimaExch96,Ptemp2005,Ptemp2020}. By exchanging temperatures, it is possible for the search to explore the feasible region further. Indeed, higher temperatures increase the probability of accepting a non-improving move, thus offering a broader coverage of the search space. This temperature exchange is known as parallel tempering. 

We use two instances of Parallel Tempering, each with 32 replicas (searches), to utilize all 64 available cores. Each replica begins with one of 32 different starting temperatures $\{T_1,T_2, \ldots, T_{32} \}$. For convenience, these temperatures are sorted in ascending order (i.e., $T_1 < T_2 < \ldots <T_{32}$) to form what some authors call a ``temperature ladder'' (e.g., \cite{BaghBoltz2020,BaghKHot2021}). While these temperatures remain constant over the entire span of the search, they are swapped between searches according to the pairwise exchange acceptance probability (EAP). A search $S_i$ with temperature $T_i$ can exchange temperatures with the search $S_{i+1}$ having the temperature one step above it on the temperature ladder, $T_{i+1}$, according to this probability: ($E_i$ is the energy function of search $i$)
\[
\text{EAP} = \min \left\{ 1, \exp{\left( \left( \frac{1}{T_i} - \frac{1}{T_{i+1}} \right)  \left( E_i - E_{i+1} \right) \right) } \right\} \, .
\] 

\section{Numerical experiments and results}
We test all three mathematical formulations, quadratic distance minimization (QP), $K$-medoids ($K$-med) and Louvain, using 12 different synthetic graphs of varying sizes and clustering difficulty. For the quadratic distance minimization and $K$-medoids formulations, we also compare solution quality and times between a leading commercial solver, Gurobi, and our two Boltzmann heuristics. Both Boltzmann machines and Gurobi were run on a 64-core/128 thread machine with SIMD instructions, with all cores available to all three solvers. Meanwhile, we use the single-core Louvain implementation of Aynaud \cite{python_louvain}, which we run on the same machine.

Our test graphs are described in Section~\ref{grahChar}. Numerical results are presented in Section~\ref{PPMres} and Section~\ref{SBMres}. The easier cases of smaller planted partition graphs are presented in Section~\ref{PPMres}. These results include trials with the Boltzmann heuristics, with Gurobi and with the Louvain heuristic. We also conduct tests using larger more complex graph structures, which are presented in Section~\ref{SBMres}. These results include trials with the Boltzmann heuristics, the Louvain heuristic and only one experiment using Gurobi. Gurobi did not return meaningful results and became unresponsive (``crashing'') after three hours of run time in the QP formulation case. It did, however, return adequate results in the case of the $K$-medoids formulation after roughly one hour of run time, before prematurely terminating (``crashing'') soon after.  We end our tests with an illustrative case-study using the famous {\it United States College Football Division IA 2000 season} graph (football graph) \cite{GirvNew2002}. This graph with known cluster structure has often been used as a ``ground-truth'' benchmark in graph clustering studies. 

\subsection{Limitation: the number of clusters}
In reviewing the numerical results, it is important to consider the pivotal importance of determining the number of clusters that best summarizes the data. As described above, the Louvain algorithm does not require the number of clusters as an input parameter. In contrast, both the QP and $K$-medoids techniques do require this input parameter. Naturally, this difference must be considered, as it represents a limitation to any conclusion based on the comparisons shown in this article.

It is also important to note that determining the number of clusters that best suits a data set is an open problem in the clustering literature. Indeed, several (general and graph-specific) common clustering techniques (e.g., $K$-means) require the number of clusters as an input parameter. As previously noted, it has been observed, in the case of graph clustering, that techniques which require the number of clusters as input tend to perform better than those which do not \cite{guideFortunato16}. This gain in performance is somewhat intuitive, given these techniques benefit from additional information.

\subsection{Synthetic graph test scenarios} \label{grahChar}
As described earlier, we conduct our experiments on 12 different synthetic graphs, with known cluster memberships. These graphs are generated using the NetworkX Python library \cite{NetworkX2008}. Our graphs are generated using two different generative models, with the use of the planted partition model (PPM) and with the stochastic block model (SBM). 

For the first set of experiments, we generate three small PPM graphs containing five clusters of 50 vertices each, for a total of 250 vertices. These graphs are generated using intra-/inter-cluster edge probabilities $(P_{\text{intra}}/P_{\text{inter}})$ of 0.9/0.1, 0.85/0.15, 0.8/0.2. These quantities also correspond to the mean intra-/inter-cluster densities of the generative model (within a margin of $10^{-3}$). Indeed, as described in Setion~\ref{Kappas}, mean intra-/inter-cluster density are empirical estimates of intra-/inter-cluster edge probability.

The second set of experiments is conducted on larger more complex SBM graph structures, each containing 5,266 vertices. This generative model allows for varying cluster sizes and varying intra-/inter-cluster edge probabilities $(P_{\text{intra}}/P_{\text{inter}})$. While we keep intra-/inter-cluster edge probabilities equal across culsters/cluster pairs, we vary cluster sizes. In these experiments, we generate graphs with intra-cluster edge probability $(P_{\text{intra}})$ of 0.8, 0.85 and 0.9. For each of those levels of intra-cluster edge probability, we create graphs with inter-cluster edge probability $(P_{\text{inter}})$ of 0.05, 0.075, 0.1. Of course, here too, these quantities correspond to the mean intra-/inter-cluster densities of the generative model (within a margin of $10^{-3}$). Clusters sizes vary between 35 and 200 vertices. This large cluster size variability along with a larger number of vertices complicate the clustering problem. Graph details are summarized in Table~\ref{graphchars}.

In summary, all of our synthetic tests are conducted with graphs that are known to be clusterable. Specifically, they are graphs whose cluster (community) structure is known in advance. Reductions in intra-cluster edge probability, increases in inter-cluster edge probability and variations in cluster sizes are used to introduce noise and complicate the cluster assignment process. Similarly, increases in graph sizes (number of vertices) are meant to increase computational challenge.

\begin{table}[]
	\centering
	\caption{Graph generative model details} \label{graphchars}
	\begin{tabular}{cccccccc}
			\hline
			Graph ID & $P_{\text{intra}}$ & $P_{\text{inter}}$ & Num Clusters & Cluster sizes & Num vertices & Num edges & Gen Model \\
			\hline 
			G1       & 0.9             & 0.1             & 5            & 50            & 250	& 8,069 	& PPM       \\
			G2       & 0.85            & 0.15            & 5            & 50            & 250   & 8,955  	& PPM       \\
			G3       & 0.8             & 0.2             & 5            & 50            & 250   & 9,955  	& PPM       \\
			G4       & 0.9             & 0.05            & 50           & {[}35,200{]}  & 5,266 & 981,435   & SBM       \\
			G5       & 0.9             & 0.075           & 50           & {[}35,200{]}  & 5,266 & 1,319,457 & SBM       \\
			G6       & 0.9             & 0.1             & 50           & {[}35,200{]}  & 5,266 & 1,654,464 & SBM       \\
			G7       & 0.85            & 0.05            & 50           & {[}35,200{]}  & 5,266 & 962,657   & SBM       \\
			G8       & 0.85            & 0.075           & 50           & {[}35,200{]}  & 5,266 & 1,301,791 & SBM       \\
			G9       & 0.85            & 0.1             & 50           & {[}35,200{]}  & 5,266 & 1,640,511 & SBM       \\
			G10      & 0.8             & 0.05            & 50           & {[}35,200{]}  & 5,266 & 945,192   & SBM       \\
			G11      & 0.8             & 0.075           & 50           & {[}35,200{]}  & 5,266 & 1,283,853 & SBM       \\
			G12      & 0.8             & 0.1             & 50           & {[}35,200{]}  & 5,266 & 1,621,210 & SBM      \\
			\hline 
		\end{tabular}
	\end{table}

\subsection{PPM results} \label{PPMres}
In Tables~\ref{QDBM}-~\ref{LouBM}, graph characteristics are displayed in the first three rows. 
$P_{\text{intra}}$ denotes the intra-cluster edge probability in the generative model and inter-cluster edge probability is denoted as $P_{\text{inter}}$. Each graph's density is denoted as $\mathcal{K}$. Clustering results appear in the lower portion of the tables.
\begin{table}[]
	\centering
	\caption{Quadratic distance minimization (QP Boltzmann)} \label{QDBM}
	\begin{tabular}{|c|l|ccc|}
		\hline
		& {Graph ID} & G1 & G2 & G3 \\
		\hline
		\multirow{3}{*}{\begin{turn}{90} Graph \end{turn}}   & $P_{\text{inter}}$  & 0.10  & 0.15 & 0.20  \\
		&$P_{\text{intra}}$  & 0.90  & 0.85 & 0.80  \\
		& $\mathcal{K}$        & 0.26 & 0.29 & 0.32 \\
		\hline
		\multirow{6}{*}{\begin{turn}{90} Results \end{turn}} & $\bar{\mathcal{K}}_{\text{inter}}$   & 0.10 & 0.15 & 0.20 \\
		&$\bar{\mathcal{K}}_{\text{intra}}$   & 0.90 & 0.85 & 0.80 \\
		& Time to sol (s)& 0.001 & 0.001 & 0.001 \\
		& $\bar{\mathcal{K}}_{\text{inter}} < \mathcal{K}$ & Y    & Y    & Y    \\
		& $\mathcal{K} < \bar{\mathcal{K}}_{\text{intra}}$ & Y    & Y    & Y   \\
		& Modularity & 0.48 & 0.38 & 0.29 \\
		\hline
	\end{tabular}
\end{table}

\begin{table}[]
	\centering
	\caption{Quadratic distance minimization (QP Gurobi)} \label{QDGB}
	\begin{tabular}{|c|l|ccc|}
		\hline
		& {Graph ID} & G1 & G2 & G3 \\
		\hline
		\multirow{3}{*}{\begin{turn}{90} Graph \end{turn}}   & $P_{\text{inter}}$ & 0.1  & 0.15 & 0.2  \\
		& $P_{\text{intra}}$ & 0.9  & 0.85 & 0.8  \\
		& $\mathcal{K}$        & 0.26 & 0.29 & 0.32 \\
		\hline
		\multirow{6}{*}{\begin{turn}{90} Results \end{turn}} & $\bar{\mathcal{K}}_{\text{inter}}$   & 0.10 & 0.15 & 0.20 \\
		&$\bar{\mathcal{K}}_{\text{intra}}$   & 0.90 & 0.85 & 0.80 \\
		& Time to sol (s)& 3.64 & 1.45 & 1.48 \\
		& $\bar{\mathcal{K}}_{\text{inter}} < \mathcal{K}$ & Y    & Y    & Y    \\
		& $\mathcal{K} < \bar{\mathcal{K}}_{\text{intra}}$ & Y    & Y    & Y   \\
		& Modularity & 0.48 & 0.38 & 0.29 \\
		\hline
	\end{tabular}
\end{table}

\begin{table}[]
	\centering
	\caption{$K$-medoids ($K$-med Boltzmann)} \label{KMBM}
	\begin{tabular}{|c|l|ccc|}
		\hline
		& {Graph ID} & G1 & G2 & G3 \\
		\hline
		\multirow{3}{*}{\begin{turn}{90} Graph \end{turn}} & $P_{\text{inter}}$ & 0.1  & 0.15 & 0.2  \\
		& $P_{\text{intra}}$ & 0.9  & 0.85 & 0.8 \\
		& $\mathcal{K}$        & 0.26 & 0.29 & 0.32 \\
		\hline
		\multirow{6}{*}{\begin{turn}{90} Results \end{turn}} & $\bar{\mathcal{K}}_{\text{inter}}$   & 0.10 & 0.15 & 0.20 \\
		&$\bar{\mathcal{K}}_{\text{intra}}$   & 0.90 & 0.85 & 0.80 \\
		& Time to sol (s)& 0.000 & 0.001 & 0.001 \\
		& $\bar{\mathcal{K}}_{\text{inter}} < \mathcal{K}$ & Y    & Y    & Y    \\
		& $\mathcal{K} < \bar{\mathcal{K}}_{\text{intra}}$ & Y    & Y    & Y   \\
		& Modularity & 0.48 & 0.38 & 0.29 \\
		\hline
	\end{tabular}
\end{table}

\begin{table}[]
	\centering
	\caption{$K$-medoids ($K$-med Gurobi)} \label{KMGB}
	\begin{tabular}{|c|l|ccc|}
		\hline
		& {Graph ID} & G1 & G2 & G3 \\
		\hline
		\multirow{3}{*}{\begin{turn}{90} Graph \end{turn}} & $P_{\text{inter}}$ & 0.1  & 0.15 & 0.2  \\ 
		& $P_{\text{intra}}$ & 0.9  & 0.85 & 0.8  \\
		& $\mathcal{K}$        & 0.26 & 0.29 & 0.32 \\
		\hline
		\multirow{6}{*}{\begin{turn}{90} Results \end{turn}} & $\bar{\mathcal{K}}_{\text{inter}}$   & 0.10 & 0.15 & 0.20 \\
		&$\bar{\mathcal{K}}_{\text{intra} }$   & 0.90 & 0.85 & 0.80 \\
		& Time to sol (s)  & 600 & 600 & 600 \\
		& $\bar{\mathcal{K}}_{\text{inter}} < \mathcal{K}$ & Y    & Y    & Y    \\
		& $\mathcal{K} < \bar{\mathcal{K}}_{\text{intra}}$ & Y    & Y    & Y   \\
		& Modularity & 0.48 & 0.38 & 0.29 \\
		\hline
	\end{tabular}
\end{table}

\begin{table}[]
	\centering
	\caption{Modularity maximization (Louvain)} \label{LouBM}
	\begin{tabular}{|c|l|ccc|}
		\hline
		& {Graph ID} & G1 & G2 & G3 \\
		\hline
		\multirow{3}{*}{\begin{turn}{90} Graph \end{turn}} & $P_{\text{inter}}$ & 0.1  & 0.15 & 0.2  \\
		& $P_{\text{intra}}$  & 0.9  & 0.85 & 0.8  \\
		& $\mathcal{K}$        & 0.26 & 0.29 & 0.32 \\
		\hline
		\multirow{7}{*}{\begin{turn}{90} Results \end{turn}} & $\bar{\mathcal{K}}_{\text{inter}}$   & 0.10 & 0.15 & 0.20 \\
		&$\bar{\mathcal{K}}_{\text{intra}}$   & 0.90 & 0.85 & 0.80 \\
		& Time to sol (s)& 0.047 & 0.062 & 0.056 \\
		& Clusters identified & 5 & 5 & 5 \\ 
		& $\bar{\mathcal{K}}_{\text{inter}} < \mathcal{K}$ & Y    & Y    & Y    \\
		& $\mathcal{K} < \bar{\mathcal{K}}_{\text{intra}}$ & Y    & Y    & Y   \\
		& Modularity & 0.48 & 0.38 & 0.29 \\
		\hline
	\end{tabular}
\end{table}

We immediately note that all four clustering formulation-solution technique combinations yield the same results, as shown in Tables~\ref{QDBM} to~\ref{LouBM}. In all three cases (G1-G3), they recover the generative model exactly. In fact, the Louvain method even recovers the exact number of clusters in the generative model.

The only distinguishing results in these experiment are the times required to obtain a solution. By far, the fastest convergence was obtained with a $K$-medoids formulation solved using a Boltzmann machine. The Louvain method, known to be very fast, was the second fastest to converge. The QP formulation solved on a Boltzmann machine was third. By far the slowest convergence was observed in the case of $K$-medoids solved with Gurobi. The QP formulation solved using the Gurobi solver was faster than in the $K$-medoids case, but orders of magnitude slower than the Boltzmann machine or Louvain.

\subsection{SBM results} \label{SBMres}
Graph characteristics and numerical results are reported in Tables~\ref{QSBM} to \ref{LouSBM}. Here too, graph characteristics are displayed in the first three rows. Again, $P_{\text{intra}}$ denotes the intra-cluster edge probability, inter-cluster edge probability is denoted as $P_{\text{inter}}$ and the graph's density as $\mathcal{K}$. We report Boltzmann experiment results obtained after 10 minutes of run time, while noting convergence had not been achieved. As mentioned previously, Gurobi failed to return meaningful results after three hours, in the case of the QP formulation. It did, however, return adequate results in the case of the $K$-medoids formulation, after roughly an hour of run time, before exiting prematurely. The Louvain algorithm converged in less than 21.5 seconds, in all instances. 

\begin{table}[]
	\centering
	\caption{Quadratic distance minimization (QP Boltzmann, 10 min run ime)} \label{QSBM}
	\begin{tabular}{|c|l|ccccccccc|}
		\hline
		& {Graph ID} & G4    & G5    & G6    & G7    & G8    & G9    & G10   & G11   & G12   \\ 
		\hline
		\multirow{3}{*}{\begin{turn}{90} Graph \end{turn}}   
		& $P_{\text{inter}}$               & 0.05 & 0.075 & 0.10 & 0.05  & 0.075 & 0.1 & 0.05 & 0.075 & 0.1 \\                  
		& $P_{\text{intra}}$               & 0.9   & 0.9   & 0.9   & 0.85  & 0.85  & 0.85  & 0.8   & 0.8   & 0.8   \\
		& $\mathcal{K}$                    & 0.07 & 0.10 & 0.12 & 0.07 & 0.09 & 0.12 & 0.07 & 0.09 & 0.12 \\
		\hline 
		\multirow{5}{*}{\begin{turn}{90} Results \end{turn}} & $\bar{\mathcal{K}}_{\text{inter}}$        & 0.06 & 0.08 & 0.11 & 0.06 & 0.08 & 0.11 & 0.06 & 0.08 & 0.11 \\
		& $\bar{\mathcal{K}}_{\text{intra}}$        & 0.76 & 0.75 & 0.75 & 0.71 & 0.71 & 0.71 & 0.67 & 0.67 & 0.67 \\
		& $\bar{\mathcal{K}}_{\text{inter}} < \mathcal{K}$ & Y     & Y     & Y     & Y     & Y     & Y     & Y     & Y     & Y     \\
		& $\mathcal{K} < \bar{\mathcal{K}}_{\text{intra}}$ & Y     & Y     & Y     & Y     & Y     & Y     & Y     & Y     & Y    \\
		& Modularity & 0.20 & 0.14 & 0.11 & 0.19 & 0.13 & 0.10 & 0.18 & 0.12 & 0.09 \\
		\hline 
	\end{tabular}
\end{table}

\begin{table}[]
	\centering
	\caption{$K$-medoids Gurobi ($K$-med Gurobi, $\sim 1$ hour run time)} \label{KMSBMGB}
	\begin{tabular}{|c|l|ccccccccc|}
		\hline 
		& Graph ID             & G4    & G5    & G6    & G7    & G8    & G9    & G10   & G11   & G12   \\
		\hline 
		\multirow{3}{*}{\begin{turn}{90} Graph \end{turn}}   
		& $P_{\text{inter}}$       & 0.05  & 0.075 & 0.10   & 0.05  & 0.075 & 0.1   & 0.05  & 0.075 & 0.1   \\ 
		& $P_{\text{intra}}$     & 0.9   & 0.9   & 0.9   & 0.85  & 0.85  & 0.85  & 0.8   & 0.8   & 0.8   \\
		& $\mathcal{K}$                    & 0.07 & 0.10 & 0.12 & 0.07 & 0.09 & 0.12 & 0.07 & 0.09 & 0.12 \\
		\hline 
		\multirow{5}{*}{\begin{turn}{90} Results \end{turn}} 
		& $\bar{\mathcal{K}}_{\text{inter}}$        & 0.05 & 0.08 & 0.11 & 0.05 & 0.08 & 0.11 & 0.05 & 0.08 & 0.11 \\
		& $\bar{\mathcal{K}}_{\text{intra}}$        & 0.83 & 0.58 & 0.41 & 0.76 & 0.56 & 0.44 & 0.66 & 0.53 & 0.26 \\
		& $\bar{\mathcal{K}}_{\text{inter}} < \mathcal{K}$ & Y     & Y     & Y     & Y     & Y     & Y     & Y     & Y     & Y     \\
		& $\mathcal{K} < \bar{\mathcal{K}}_{\text{intra}}$ & Y     & Y     & Y     & Y     & Y     & Y     & Y     & Y     & Y    \\
		& Modularity & 0.28 & 0.15 & 0.08 & 0.26 & 0.15 & 0.09 & 0.24 & 0.14 & 0.04 \\
		\hline 
	\end{tabular}
\end{table}

\begin{table}[]
	\centering
	\caption{$K$-medoids ($K$-med Boltzmann, 10 min run time)} \label{KMSBMBM}
	\begin{tabular}{|c|l|ccccccccc|}
		\hline 
		& Graph ID             & G4    & G5    & G6    & G7    & G8    & G9    & G10   & G11   & G12   \\
		\hline 
		\multirow{3}{*}{\begin{turn}{90} Graph \end{turn}}   
		& $P_{\text{inter}}$       & 0.05  & 0.075 & 0.10   & 0.05  & 0.075 & 0.1   & 0.05  & 0.075 & 0.1   \\ 
		& $P_{\text{intra}}$     & 0.9   & 0.9   & 0.9   & 0.85  & 0.85  & 0.85  & 0.8   & 0.8   & 0.8   \\
		& $\mathcal{K}$                    & 0.07 & 0.10 & 0.12 & 0.07 & 0.09 & 0.12 & 0.07 & 0.09 & 0.12 \\
		\hline 
		\multirow{5}{*}{\begin{turn}{90} Results \end{turn}} 
		& $\bar{\mathcal{K}}_{\text{inter}}$        & 0.05 & 0.08 & 0.10 & 0.05 & 0.08 & 0.11 & 0.05 & 0.08 & 0.10 \\
		& $\bar{\mathcal{K}}_{\text{intra}}$        & 0.90 & 0.84 & 0.77 & 0.85 & 0.80 & 0.61 & 0.77 & 0.66 & 0.65 \\
		& $\bar{\mathcal{K}}_{\text{inter}} < \mathcal{K}$ & Y     & Y     & Y     & Y     & Y     & Y     & Y     & Y     & Y     \\
		& $\mathcal{K} < \bar{\mathcal{K}}_{\text{intra}}$ & Y     & Y     & Y     & Y     & Y     & Y     & Y     & Y     & Y    \\
		& Modularity & 0.28 & 0.20 & 0.14 & 0.27 & 0.19 & 0.12 & 0.26 & 0.17 & 0.13 \\
		\hline 
	\end{tabular}
\end{table}

\begin{table}[]
	\centering
	\caption{Louvain} \label{LouSBM}
	\begin{tabular}{|c|l|lllllllll|}
		\hline 
		& Graph ID             & G4    & G5    & G6    & G7    & G8    & G9    & G10   & G11   & G12   \\
		\hline
		\multirow{3}{*}{\begin{turn}{90} Graph \end{turn}}   
		& $P_{\text{inter}}$               & 0.05  & 0.075 & 0.1   & 0.05  & 0.075 & 0.1   & 0.05  & 0.075 & 0.1   \\ 
		& $P_{\text{intra}}$                & 0.9   & 0.9   & 0.9   & 0.85  & 0.85  & 0.85  & 0.8   & 0.8   & 0.8   \\
		& $\mathcal{K}$                     & 0.07 & 0.10 & 0.12 & 0.07 & 0.09 & 0.12 & 0.07 & 0.09 & 0.12 \\
		\hline 
		\multirow{5}{*}{\begin{turn}{90} Results \end{turn}} & $\bar{\mathcal{K}}_{\text{inter}}$        & 0.05 & 0.08 & 0.11 & 0.05 & 0.08 & 0.11 & 0.05 & 0.08 & 0.11 \\
		& $\bar{\mathcal{K}}_{\text{intra}}$       & 0.56 & 0.63 & 0.62 & 0.71 & 0.59 & 0.63 & 0.60 & 0.60 & 0.51 \\
		& Clusters identified & 21	&22	&18	&30	&22	&18	&22	&25	&17 \\
		& $\bar{\mathcal{K}}_{\text{inter}} < \mathcal{K}$ & Y     & Y     & Y     & Y     & Y     & Y     & Y     & Y     & Y     \\
		& $\mathcal{K} < \bar{\mathcal{K}}_{\text{intra}}$ & Y     & Y     & Y     & Y     & Y     & Y     & Y     & Y     & Y   \\
		& Modularity & 0.27 & 0.20 & 0.15 & 0.27 & 0.19 & 0.14 & 0.25 & 0.18 & 0.13 \\
		\hline  
	\end{tabular}
\end{table}

Three notable results appear in this set of experiments. First, we note that all three formulations, QP, $K$-medoids and Louvain lead to arguably good clustering, regardless of the numerical solution technique. In all cases, the inequalities $\bar{\mathcal{K}}_{\text{inter}} < \mathcal{K} < \bar{\mathcal{K}}_{\text{intra}}$ hold. Second, we note that clustering quality of the distance minimization, $K$-medoids are roughly equivalent or superior to the clustering quality obtained with the Louvain method, depending of the numerical solution technique used (Gurobi or BM). Finally, we note that Louvain provides, by far, the fastest solutions (21.5 seconds vs. 600+).

A closer examination of intra-cluster densities, shown in Table~\ref{intras}, highlights the differences in solution quality. First, we immediately note that the intra-cluster densities identified by the Louvain method remain stuck between 60 and 70\%. Regardless of the underlying graph structure, they seem to vary randomly within that interval. In contrast, the intra-cluster densities identified by both the distance minimization and $K$-medoids formulations decrease with intra-cluster edge probabilities and with increases in inter-cluster edge probabilities. We also note that while BM $K$-medoids offers higher, more accurate, intra-cluster densities than the (BM) QP formulation, in most cases, it also appears more sensitive to inter-cluster edge probabilitiy. 

Our results also underscore the strength of our BM solver. Indeed, even with the same mathematical formulation ($K$-med) and the benefit of a much longer run time, Gurobi yields sparser clusters than our BM. These results can be seen in Tables~\ref{KMSBMGB},~\ref{KMSBMBM},~\ref{intras}. More importantly, our one-hot BM was also able to obtain competitive results for the QP formulation in the cases of the larger more complex SBM graphs. In those same cases, Gurobi not only failed to converge after several hours of run time, it became unresponsive .

Finally, our results highlight the disconnection between modularity and intra-cluster density. While the Louvain heuristic, predictably, yields the highest modularity levels, it fails to obtain the densest clusters. Our results also show that, in many instances, clusterings with lower modularity are in fact denser. For example, in Table~\ref{QSBM} we see a modularity of 0.10 and a mean intra-cluster density $\bar{\mathcal{K}}_{\text{intra}} = 0.71$ obtained with the (BM) QP formulation-solver combination, in the case of graph $G9$. Meanwhile, for the same graph, the Louvain technique yields a higher modularity (0.14), but a much lower mean intra-cluster density $\bar{\mathcal{K}}_{\text{intra}} = 0.60$. These results are consistent with previous experiments comparing modularity and density \cite{PMEtAlWAW18,PMEtAlOUP20}.

\begin{table}[]
	\centering
	\caption{Side by side comparisons of $\bar{\mathcal{K}}_{\text{intra}}$} \label{intras}
	\begin{tabular}{cccc|cccc}
		\hline 
		&&  && \multicolumn{3}{c}{Mathematical formulation} \\
		\multicolumn{4}{c|}{Graph characteristics} & Louvain & QP BM & $K$-med Gurobi & $K$-med BM \\
		\hline 
		Graph ID & $P_{\text{inter}}$       & $P_{\text{intra}}$       & $\mathcal{K}$     & $\bar{\mathcal{K}}_{\text{intra}}$ & $\bar{\mathcal{K}}_{\text{intra}}$ & $\bar{\mathcal{K}}_{\text{intra}}$  & $\bar{\mathcal{K}}_{\text{intra}}$\\
		G4 & 0.05  & 0.9  & 0.071 & 0.56 & 0.76 & 0.83 &{\bf 0.90} \\
		G5 & 0.075  & 0.9 & 0.095 & 0.63 & 0.75 & 0.58 & {\bf 0.84} \\
		G6 & 0.1  & 0.9   & 0.119 & 0.62 & 0.75 & 0.41 & {\bf 0.77} \\ 
		G7 & 0.05 & 0.85  & 0.069 & 0.71 & 0.71 & 0.76 & {\bf 0.85} \\
		G8 & 0.075 & 0.85 & 0.094 & 0.59 & 0.71 & 0.56 & {\bf 0.80}  \\
		G9 & 0.1 & 0.85   & 0.118 & 0.63 & {\bf 0.71} & 0.44 & 0.61   \\
		G10 & 0.05 & 0.8  & 0.068 & 0.60 & 0.67 & 0.66 & {\bf 0.77} \\
		G11 & 0.075 & 0.8 & 0.093 & 0.60 & {\bf 0.67} & 0.53 &  0.66              \\
		G12 & 0.1 & 0.8   & 0.117 & 0.51 & {\bf 0.67} & 0.26 & 0.65                \\
		\hline 
		{\bf Count Best of 3} &  &&& {\bf 0} & {\bf 3} & {\bf 0} & {\bf 6} \\
		 \hline
	\end{tabular}
\end{table}

\subsection{Illustrative case study: the US College Football graph} \label{rw}
Our earlier numerical tests using synthetic graphs are designed to compare the ability of each technique and solver under study to identify densely connected subgraphs. However, we find it useful to also examine the ability of each formulation and solver to recover the cluster membership of a real-world graph's vertices with known cluster membership. For these comparisons, we use the famous {\it United States College Football Division IA 2000 season} graph (football graph) \cite{GirvNew2002}. This graph is a representation of the regular season encounters between 115 college football teams. Each team is represented by a vertex. These vertices grouped into one of twelve conferences (clusters). Edges connect teams that faced each other at least once during the regular season. Teams within a conference all face each other during regular season, while they do not necessarily face teams outside their conference during the regular season. Therefore, there are more shared connections between teams of the same conference than between teams in different conferences. Graph characteristics are provided in Table~\ref{fb graph}.

\begin{table}[]
\centering
\caption{US College Football Division IA 2000 season graph charateristics} \label{fb graph}
\begin{tabular}{|l|c|}
\hline
$\vert V \vert$  (vertices/teams)      & 115  \\
\hline
$\vert E \vert$ (edges)       & 613  \\
\hline
$\mathcal{K}$ (density)            & 0.09 \\
\hline
Clusters (conferences) & 12 \\ 
\hline
\end{tabular}
\end{table}

While this examination reveals interesting results, we find it important to highlight its limitations. The objective functions of the techniques discussed in this article are designed to yield the clusterings with densest subgraphs (QP), the most representative exemplars ($K$-med) or the clusterings with the highest modularity, at their respective optima. In accordance with the universally accepted notion that a successful graph clustering yields densely connected clusters (subgraphs), we assess clustering quality via the mean intra-cluster density of the clusters identified through each of the techniques/solvers in this study.

Unfortunately, the ground-truth cluster membership of the vertices in a real-world graph may not correspond to the clustering with the densest clusters. After all, real-world graphs are instances of typically unknown latent generative models and random noise. In many cases, it may be possible to modify the cluster assignments of a labeled real-world graph's vertices and obtain higher intra-cluster densities. For this reason, results in this section should be taken in context and not understood as a definitive ranking of the various optimization techniques and solvers. More generally, it should be noted that ground-truth graphs are not typically the best benchmarks for clustering quality assessments.

Another challenge posed by this examination is the measurement of clustering accuracy with respect to the ground-truth clustering. Here, it is not sufficient to compare clusters according to their labels, because these labels are arbitrary. For example, cluster $c_1^g$ in the ground-truth (superscript $g$) labeling may correspond exactly to cluster $c_2^a$ returned by the clustering procedure (superscript $a$). Simply comparing clusters according to their labels (e.g., comparing $c_1^g$ to $c_1^a$) is not meaningful. Instead, cluster constituents must be compared. Each ground-truth cluster must be compared to each cluster identified by a clustering procedure, to assess the similarity of cluster contents. 

Clusters, be they ground-truth or identified by an algorithm, are disjoint sets of vertices (empty set intersection). For this reason, we use Jaccard similarity function ($J$) to compare contents of ground-truth clusters ($c_i^g$) and those identified by an algorithm ($c_j^a$). We take the maximum Jaccard ($\tilde{J}$) similarity over all possible ground-truth clusters as a gauge of similarity between a cluster identified by an algorithm and its associated ground-truth benchmark:
\[
\tilde{J}(c_j^a) = \max_{c_i^g} \left\{ J(c_i^g,c_j^a) = \frac{\vert c_i^g \cap c_j^a \vert}{\vert c_i^g \cup c_j^a \vert} \right\} \, .
\] 

Each $\tilde{J}(c_j^a)$ provides a score for the similarity of a cluster as identified by clustering algorithm ($c_j^a$) and its associated benchmark. A perfect match between a cluster identified by clustering algorithm and its associated benchmark yields a value of $\tilde{J}(c_j^a) = 1$, while a complete mismatch yields a value of $\tilde{J}(c_j^a) = 0$. To obtain a graph-level view and a valid comparison, we compute the means of the $\tilde{J}$ over all clusters. Comparisons are presented in Table~\ref{fbresults}. These comparison reveal that the $K$-medoids formulation provides a better match to the ground-truth clustering.

\begin{table}[]
	\centering
	\caption{Similarity to ground-truth clusters} \label{fbresults}
	\begin{tabular}{r|cccccc}
		& Louvain & QP Gurobi & QP BM & $K$-med Gurobi & $K$-med BM \\
		\hline
		Num exact matches ($\tilde{J} = 1$) & 4 & 3 & 3 & 6 & 6 \\
		Mean  $\tilde{J}$ & 0.72    & 0.78      & 0.81  & 0.84         & 0.83    \\
		\hline
	\end{tabular}
\end{table}

\section{Conclusion}
In this article, we have successfully adapted the quadratic distance minimization (QP) and quadratic $K$-medoids formulations to the graph clustering problem. These formulations provide better results than the well established ``state of the art'' Louvain method. We also illustrate the value of a Boltzmann heuristic in solving NP-hard combinatorial optimization problems.

Future work will focus on additional comparisons, $K$-medoid parameter tuning, clustering with overlapping clusters and scalability. The Louvain method has been applied to very large scale commercial problems. Any alternative must be applicable to similarly large scale problems. Finally, because both our mathematical formulations rely heavily on vertex-vertex distance, we also intend to pursue further examinations of the topic.

\section*{Acknowledgements}
The authors would like to thank Fujitsu Limited and Fujitsu Consulting (Canada) Inc.~for providing financial support.

\section*{Availability of data and materials}
Most of the data used in our experiments was generated using the NetworkX library's planted partition and stochastic block model generators \cite{NetworkX2008}. All parameters required to reproduce our data are in Table~\ref{graphchars}.

The football graph used in Section~\ref{rw} was obtained from Mark Newman's web site. http://www-personal.umich.edu/~mejn/netdata/football.zip

\section*{Competing interests} 
The authors declare that they have no competing interests.

\section*{Funding}
The authors received financial support from Fujitsu Limited and Fujitsu Consulting (Canada) Incorporated.

\bibliographystyle{spmpsci}      
\bibliography{myBib}

\end{document}